%% file: main.tex
\documentclass[letterpaper,10pt,conference]{ieeeconf}
\IEEEoverridecommandlockouts
\usepackage{amsmath,amssymb}
\usepackage{graphicx}
\usepackage{booktabs}
\usepackage[acronym]{glossaries}
\usepackage{makecell}
\usepackage{tabularx}
\usepackage{array}
\usepackage{xcolor}
\usepackage{ragged2e}
\usepackage{float}
\usepackage{pgfplots}
\pgfplotsset{compat=1.18}
\newcolumntype{L}{>{\RaggedRight\arraybackslash}X}

\input{./tex/preamble}

\title{\LARGE \bf
Training Humans to Teach Robots: Upskilling, Transfer and Retention of Visuomotor Demonstration Skills}

\author{Endong Sun$^{1}$, Yuqing Zhu and Matthew Howard%
\thanks{$^{1}$Endong Sun ({\tt\small endong.sun@kcl.ac.uk}), Yuqing Zhu and Matthew Howard are with the Centre for Robotics Research, Department of Engineering, King's College London, UK.}}

\begin{document}
\raggedbottom
\maketitle

\begin{abstract}
Teleoperation and \AR\ interfaces make demonstrations easier to collect, but not necessarily easier for robots to learn from.
This paper introduces \DEMT, which evaluates successful demonstrations through robot deployment and converts into \AR\ guidance for human training.
In simulation, all demonstrations complete the task, yet variation in path, contact orientation, or timing lowers robot deployment success by 70 percentage points on average.
A controlled study with 20 novices shows that training improves robot learning even after guidance is removed.
Policies learned from trained participants' demonstrations achieve success rates 63 percentage points higher on average on block manipulation and 55 percentage points higher on a longer-horizon task than those learned from untrained participants.
Two months later, without refresher training, trained participants retain a teaching advantage on new tasks: robot success rates are higher by 24 percentage points on obstacle avoidance with multiple valid solutions and by 39 percentage points on average on unfamiliar fruits.
These findings establish human teaching skill as a resource that can be trained, transferred, and retained, making demonstrator training a complementary route to better robot learning without increasing the demonstration budget.

\end{abstract}

\glsresetall%
\section{Introduction}
\label{s:introduction}

Recent visuomotor \LfD\ methods, including Diffusion Policy and the Action Chunking Transformer, have enabled substantial progress in robot manipulation~\cite{chi2025diffusion,zhao2023learning}.
The performance of these policies, however, depends not only on model architecture but also on the quantity and quality of the demonstrations used for training.
In-the-wild and low-cost teleoperation systems have made large-scale collection increasingly accessible, including to novice users~\cite{chi2024universal,iyer2024open,wang2024dexcap,fu2025mobilealoha,chen2025arcap,yu2025egomi}.
Scalable collection reduces the cost of obtaining demonstrations, but does not ensure that the collected data are useful to a particular robot learner.

Demonstration quality is often associated with successful task completion.
Yet human demonstrators may rely on perceptual, tactile, and corrective capabilities that differ from those available to the robot learner.
The robot must instead infer an observation-to-action mapping under its own embodiment, perception, controller, policy class, and finite data budget~\cite{sena2020quantifying,belkhale2023dataquality,sakr2024consistency,hejna2025deminf,kumar2025pacer}.
Consequently, datasets composed entirely of task-successful demonstrations can induce policies with different deployment performance.

This work investigates demonstrator training as a means of improving robot learning under a fixed, empirically sufficient demonstration budget.
To this end, \DEMT\ is introduced as a pre-collection framework that selects human-training targets based on the deployment performance of a fixed robot learner.
For visuomotor manipulation, candidate targets specify end-effector path, contact orientation, and motion pacing.
The selected target is then converted into \AR\ guidance comprising corridor, grasp, and speed cues for novice training curriculum.

The evaluation combines controlled simulation with a study of 20 novices and a two-month follow-up.
Simulation compares task-successful demonstration structures, while the human study tests teaching improvement, immediate transfer, and retained transfer without guidance during evaluation data collection.

Overall, this paper addresses two central questions: (a) Do demonstrations that successfully complete a task necessarily lead to reliable robot policies? (b) Can training develop novice teaching skills that improve subsequent robot learning, transfer beyond the training curriculum, and persist without refresher training?

In addressing these questions, this paper makes two contributions:
\begin{enumerate}[leftmargin=*,label=(\roman*)]
    \item \DEMT\ provides a deployment-based criterion for selecting human-training targets under a fixed robot learner and demonstration budget. Controlled simulation demonstrates that task-successful datasets can differ substantially in robot learnability.
    \item The human study provides evidence that novice robot-teaching skills can be trained, transferred beyond the guided curriculum, and retained over two months: demonstrations collected without guidance yield higher robot deployment success than those from untrained controls.
\end{enumerate}

\begin{figure*}[!t]
    \centering
    \includegraphics[width=\textwidth]{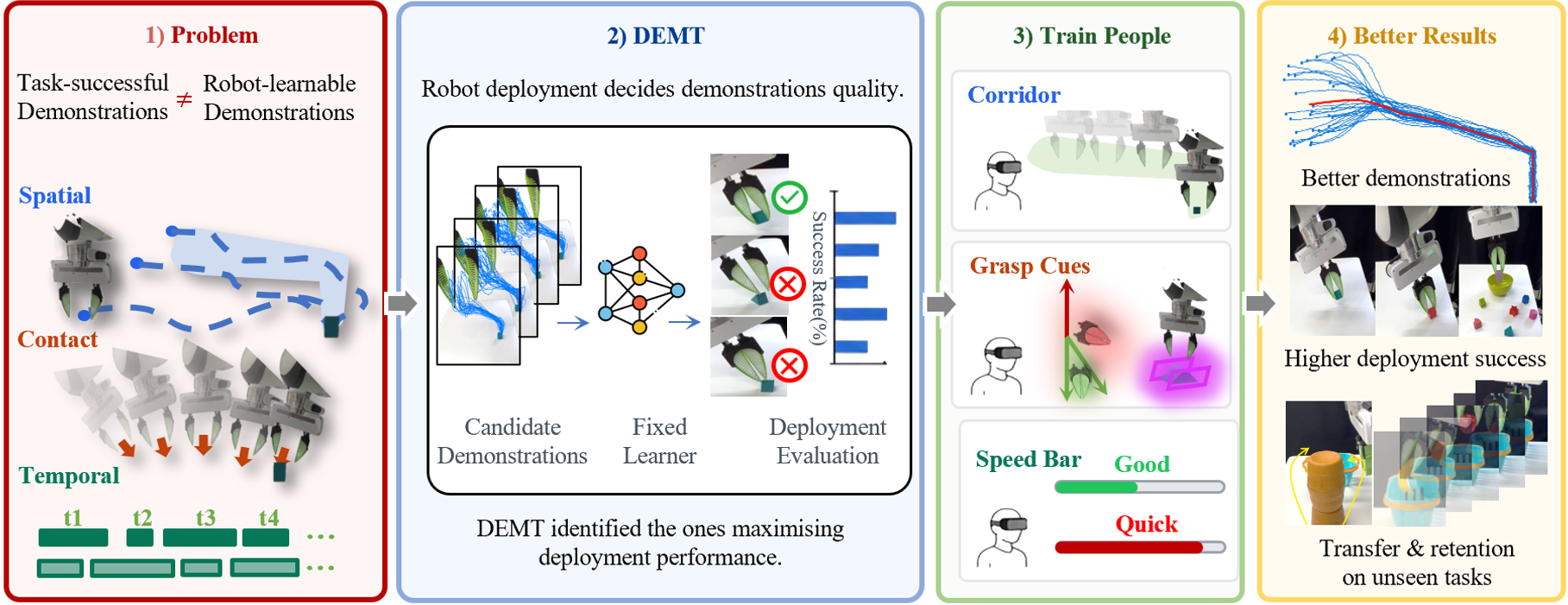}

    \caption{\DEMT\ trains people to provide demonstrations that support robot learning.
    Candidate demonstration structures are evaluated through policy deployment, and the selected structure is translated into AR training cues.
    Subsequent data collection without these cues tests immediate and two-month retained transfer of the acquired teaching skill.}
    \label{fig:framework_overview}
    \vspace{-2.5mm}
\end{figure*}

\section{Related Work}
\label{s:related_work}

\emph{Scaling visuomotor demonstration collection:}
Recent systems reduce the hardware and expertise required to gather robot demonstrations.
UMI enables portable in-the-wild collection~\cite{chi2024universal}; Mobile ALOHA extends low-cost teleoperation to mobile bimanual manipulation~\cite{fu2025mobilealoha}; and Open Teach, DexCap, and EgoMI use mixed reality, dexterous capture, or egocentric retargeting to broaden access~\cite{iyer2024open,wang2024dexcap,yu2025egomi}.
Complementing these advances in accessibility, this work studies how novices can learn to provide demonstrations that a fixed visuomotor learner can use effectively.

\emph{Learner-aware demonstration quality:}
Successful demonstrations can differ substantially in value for learning.
Prior work characterises quality through action divergence, transition diversity, consistency, predictability, or estimated influence~\cite{sakr2022quantifying,belkhale2023dataquality,sakr2024consistency,hejna2025deminf}.
Data-centric methods improve collected datasets through filtering, reweighting, selection, relabelling, or mixture optimisation~\cite{agia2025cupid,hejna2025remix,zhang2025scizor,lee2026qoq,sagheb2025counterbc}.
Machine teaching instead treats the teaching set as a learner-dependent design variable~\cite{zhu2018overview}.
\DEMT\ retains this learner-aware perspective but evaluates human-controllable structures through robot deployment before target data collection.

\emph{Guiding and training human demonstrators:}
ARCap, ARMimic, and ARMADA provide collection-time feedback related to viewpoint, feasibility, collision, or execution~\cite{chen2025arcap,walia2025armimic,nechyporenko2024armada}; other interfaces use feasibility cues, force feedback, or refinement~\cite{yin2026feasiblecap,JoyLo,10904309,electronics15081653}.
These systems support demonstrators during recording.
Pre-collection training instead uses guidance to develop teaching behaviour that can subsequently be applied independently.
Prior studies examine learner-aware human teaching and retained or transferred teaching behaviour in lower-dimensional settings~\cite{sena2020quantifying,zhu2026training,sun2025training}.
The present study extends this direction to visuomotor manipulation, evaluating the acquired skill through robot policies learned from subsequent demonstrations, both immediately and two months after training.

\section{DEMT: Framework}
\label{s:framework}

\begin{figure*}[!t]
    \centering
    \includegraphics[width=\textwidth]{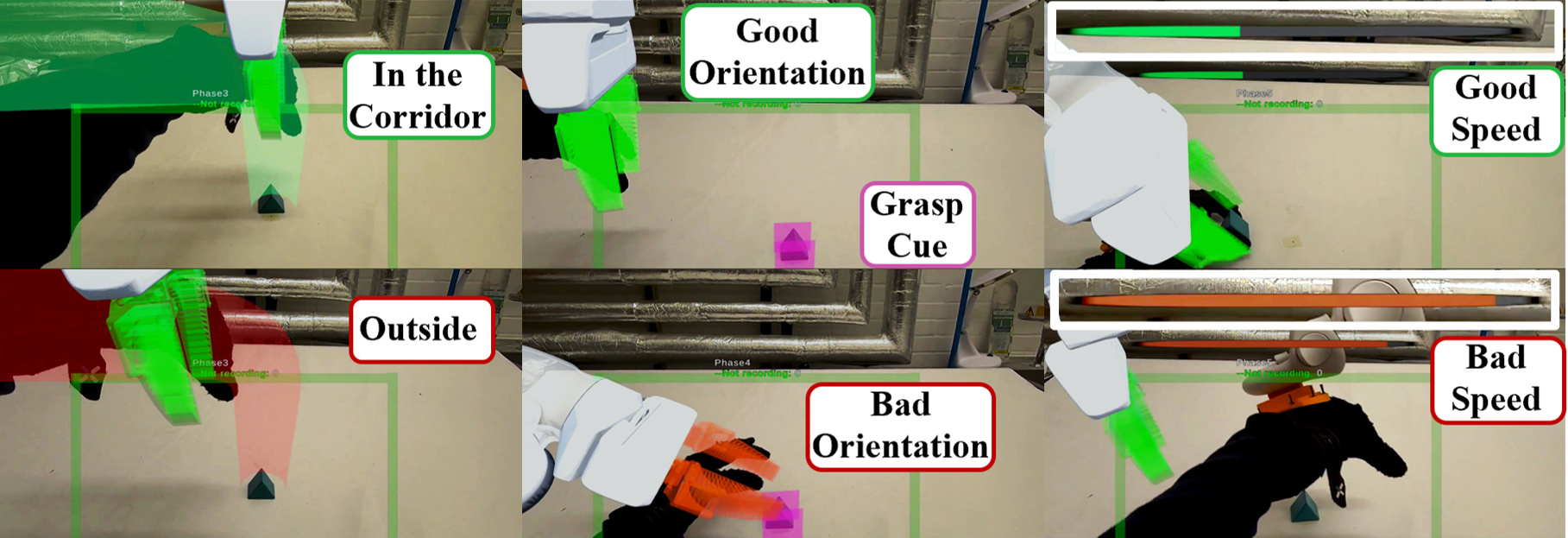}

    \caption{AR cues used to train the selected demonstration structure.
    Spatial corridor, gripper contact, and speed-bar cues provide feedback on path, grasp orientation, and motion speed, respectively.
    These training cues are removed before subsequent policy-data collection.}
    \label{fig:guidance}
    \vspace{-2.5mm}
\end{figure*}

\subsection{Problem Definition}
\label{s:problem}

This paper considers teaching visuomotor manipulation through demonstrations, using learners such as 3D Diffusion Policies~\cite{chi2025diffusion,wang2024dexcap,chen2025arcap}.
A demonstration dataset $\dataset\in\datasetspace$ contains observations ($\obsc\in\obsspace$) and corresponding actions ($\actionc\in\actionspace$). The learning algorithm $\learner(\cdot)$ maps this dataset to the learnt model
\begin{equation}
\label{e:dp_learner}
\learntmodel = \learner(\dataset),\quad \dataset = (\obsc , \actionc ).
\end{equation}

To distinguish successful task execution from effective teaching, demonstration quality must be assessed in relation to the robot learner.
Machine teaching~\cite{zhu2018overview} provides this learner-dependent perspective and can optimise teaching quality at a fixed budget, or minimise the budget needed to attain a prescribed teaching quality.
This paper adopts the first formulation: the budget is fixed at the smallest level found empirically to support reliable, high deployment performance, and teaching seeks the dataset that best induces the desired model $\targetmodel$ within that budget.
In model space, this gives the following bi-level optimisation problem:
\begin{equation*}
\begin{aligned}
\optimaldataset
&=
\arg\min_{\dataset\in\datasetspace}
\triskf(\learntmodel,\targetmodel),\\
&\mathrm{s.t.}\quad
\learntmodel=\learner(\dataset),\quad
\terrorf(\dataset)=\Nd .
\end{aligned}
\end{equation*}
The teaching risk $\triskf(\cdot)$ measures the discrepancy between the learnt and target models, while $\terrorf(\dataset)$ counts demonstrations under the fixed budget $\Nd$.
Evaluating this model-space risk requires an explicitly specified target model $\targetmodel$.
For visuomotor manipulation, the desired task outcome is easier to specify than a target neural-network parameter vector, motivating evaluation through robot deployment.

\subsection{Deployment-Evaluated Machine Teaching}

\DEMT\ therefore evaluates a dataset by the deployment performance of the policy learned from it, not by the demonstrations' completion.
It replaces model-space teaching risk with
\begin{equation}
\label{e:empirical_trf}
J(\learner(\dataset))
=
\mathbb{E}\left[\successf(\learner)\right],
\end{equation}
where \(\successf(\cdot)\) denotes the deployment protocol, including task completion, stable contact, safe motion, and absence of damaging collisions.

Let \(\mathcal{Z}\) denote the search space of all demonstration structures that can serve as training targets, and let \(\dataset_z\) denote the dataset induced by \(z\in\mathcal{Z}\).
Given a fixed learner, embodiment, sensing and action pipeline, deployment protocol, and budget, \DEMT\ selects the deployment-optimal training structure as
\begin{equation}
\label{e:structure_demt_obj}
z^\star =
\arg\max_{z\in\mathcal{Z}}
J\!\left(\learner(\dataset_z)\right)
\quad
\mathrm{s.t.}\quad
|\dataset_z|=\Nd .
\end{equation}
\subsection{Spatial--Contact--Temporal Instantiation}
\label{s:prv_instantiation}

To turn the deployment objective in Eq.~\eqref{e:structure_demt_obj} into a human-training target within the framework shown in Fig.~\ref{fig:framework_overview}, demonstration structures must be expressed in terms that people can control.
Informed by prior \LfD\ analyses of demonstration quality~\cite{sakr2022quantifying,belkhale2023dataquality,sakr2024consistency,hejna2025deminf}, this study uses the PRV teaching space
\[
\mathcal{Z}_{\mathrm{PRV}}
=
\mathcal{P}\times\mathcal{R}\times\mathcal{V},
\]

\begin{itemize}[leftmargin=*]
    \item \textbf{Spatial structure (\(\mathcal{P}\))}: the end-effector path, specifying \textit{where} the robot moves.
    \item \textbf{Contact structure (\(\mathcal{R}\))}: the contact or grasp strategy, specifying \textit{how} the robot interacts.
    \item \textbf{Temporal structure (\(\mathcal{V}\))}: the speed profile and phase timing, specifying \textit{when} critical actions occur.
\end{itemize}

Within this instantiation, the finite candidate set is manually defined as
\[
\mathcal{C}
=
\{z_0,z_{\mathcal{P}},z_{\mathcal{R}},z_{\mathcal{V}}\}
\subset
\mathcal{Z}_{\mathrm{PRV}}.
\]
Candidate \(z_0\) maintains consistent spatial, contact, and temporal structures; the remaining candidates isolate variation in one structure while holding the others fixed.
Task completion alone does not distinguish these successful demonstration structures; \DEMT\ instead compares how well the robot learns from each structure.
The training target is selected by empirical deployment performance:
\begin{equation}
\label{e:detail_demt_obj}
\hat{z}
=
\arg\max_{z\in\mathcal{C}}
\hat{J}\!\left(\learner(\dataset_z)\right)
\quad
\mathrm{s.t.}\quad
|\dataset_z|=\Nd .
\end{equation}
The selected \(\hat z\) is the empirically best candidate in \(\mathcal C\), providing a practical approximation to the broader objective over \(\mathcal Z\).

\subsection{Hand-Crafted AR Guidance}
\label{s:method_training_guidance}

Selecting an effective demonstration structure does not establish that novices can learn to demonstrate it.
This motivates the second central question: whether practising the selected structure develops teaching skill that remains useful beyond guided training.
The selected PRV structure is translated into three AR cues in the task space (Fig.~\ref{fig:guidance}), linking each training target to direct visual feedback~\cite{norman1988design}:
\begin{enumerate}[label=$g_\arabic*$,leftmargin=*]
    \item \label{i:g1}\textbf{Spatial corridor.}
    A tube-like corridor visualises the target path; motion inside is green and motion outside red.
    \item \label{i:g2}\textbf{Gripper contact cue.}
    Purple contact layers indicate an expert-selected opposing grasp region; green/red grippers indicate whether contact orientation lies within the target family.
    \item \label{i:g3}\textbf{Speed bar.}
    A task-normalised bar displays motion speed relative to a reference-based upper bound throughout the trajectory.
\end{enumerate}

\subsection{Hypotheses}

Hypothesis \(h_1\) addresses the first central question; \(h_2\)--\(h_4\) address the trainability, transfer, and retention of human teaching skill in the second.
Under a fixed learner, embodiment, sensing pipeline, deployment protocol, and budget, the following hypotheses are tested:
\begin{enumerate}[label=$h_\arabic*$,leftmargin=*]
    \item \label{i:h1}\textbf{Task success is insufficient for robot learnability:} task-successful datasets with different PRV structures can induce different deployment performance.
    \item \label{i:h2}\textbf{Robot-teaching behaviour can be improved:} novices provide more learnable demonstrations after \AR\ guidance removal.
    \item \label{i:h3}\textbf{The improved teaching behaviour transfers\footnote{Transfer refers to the human teaching skill; a new robot policy is learned for each evaluation task.}:} after training, novices apply the acquired teaching skill to new objects, related tasks, and tasks with multiple valid solutions.
    \item \label{i:h4}\textbf{The improved teaching behaviour transfer is retained:} novices retain and apply the acquired teaching skill across complex tasks two months later without refresher guidance.
\end{enumerate}

\section{Experimental Design}
\label{s:experiments}
The evaluation proceeds from a simulation($h_1$), the immediate human tranining and transfer($h_2,h_3$), to a follow-up of retained transfer($h_4$). Within each comparison, the learner, embodiment, sensing and action spaces, training procedure, and demonstration budget are held fixed.
All policies use a 3D Diffusion Policy with a PointNet encoder and 1D conditional U-Net: 2000 XYZ--RGB points, \(T_o=5\), \(T_p=20\), batch size 16, 3000 epochs, Adam at \(10^{-4}\), one fixed seed, and 100/10 DDIM training/inference steps.
The budget \(\Nd=30\) was empirically selected as the smallest budget that reliably induced high deployment performance with this learner and setup.
This budget is used per task, or per prescribed route in the bimodal experiment.
The physical experiments use a Franka Panda robot.
The sampling space for initial end-effector positions is \([0.0,0.6]\!\times\![-0.1,0.1]\!\times\![0.0,0.6]\,\mathrm{m}\); control runs at 30 Hz and replans after holding each predicted action for five steps. Across these tests, the primary outcome is empirical deployment success rate (EDSR; Sec.~\ref{s:metrics}) of policies learned from the resulting demonstrations.

\subsection{Simulation to Test the Task-Success--Learnability Gap}
\label{s:exp_sim}

\begin{figure*}[!t]
    \centering
    \includegraphics[width=\textwidth]{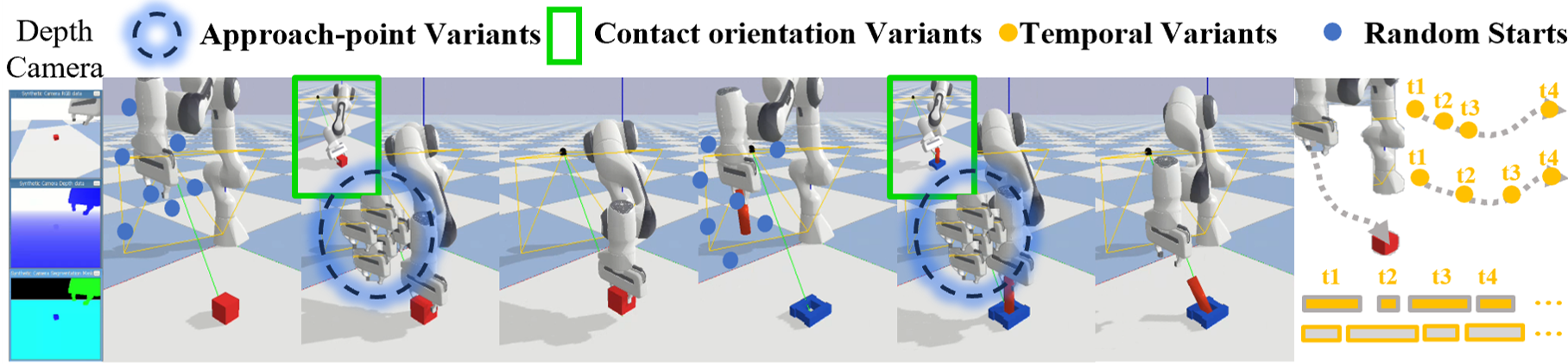}

    \caption{Finite-candidate grasping and insertion study.
    The left inset shows depth observations; blue markers denote random starts.
    Dashed blue regions, green outlines, and orange markers denote spatial, contact, and temporal variants.
    All scripted demonstrations succeed with each variant changing one PRV structure while holding the others fixed.}
    \label{fig:sim_variants}
    \vspace{-2.5mm}
\end{figure*}

To test whether task-successful demonstrations can yield different robot performance (\(h_1\)), the simulation study holds demonstration task success constant while varying demonstration structure within the finite candidate set \(\mathcal{C}\subset\mathcal{Z}_{\mathrm{PRV}}\) from Sec.~\ref{s:prv_instantiation}.
Two atomic skills on Pybullet simulator are considered: parallel-jaw grasping and peg insertion.
For each task, \(z_0\) maintains common spatial paths, contact orientations, and temporal pacing, while \(z_{\mathcal P}\), \(z_{\mathcal R}\), and \(z_{\mathcal V}\) vary one structure while holding the other two fixed. Demonstration are scripted in this simulation.
Every demonstration completes the task without collision or inverse-kinematics failure.
Each condition contains four independently generated datasets of \(\Nd=30\) demonstrations; a 3D Diffusion Policy is trained from each dataset and evaluated for \(K=10\) rollouts.
Differences in EDSR across these task-successful conditions would support \(h_1\): completing the task is not sufficient to ensure equally learnable data.
The same deployment comparison selects the training target through Eq.~\eqref{e:detail_demt_obj}, connecting this test to the subsequent human curriculum.
Figure~\ref{fig:sim_variants} illustrates the variants and their shared sensing and random-start setup.

\subsection{Immediate Human Training and Transfer}
\label{s:exp_immediate}

Having identified a learner-effective structure, the next question is whether novices can learn to provide it independently (\(h_2\)) and apply the acquired skill beyond the guided task (\(h_3\)).
From a power analysis (HHU G*Power) for detecting the effect size with a 0.1 error probability and 0.8 power, twenty novice participants are assigned to the Target and Control groups by blinded random allocation, constrained to 10 participants per group\footnote{The study was approved by 
King's College London (Approval No.~[MRSP-25/26-55721]
); all participants provided informed consent. All participants are right-handed and have no prior knowledge of robotics.}.
Both groups use the same Meta Quest 3 interface, Franka Panda setup, tasks, and number of demonstrations.
The Target group receives \AR\ guidance comprising spatial corridor, contact-orientation, and speed cues; the Control group performs matched practice without \AR\ guidance.
Each phase contains a fixed budget of 30 demonstrations.

The training curriculum comprises the following phases:
\begin{itemize}[leftmargin=*]
    \item \textbf{P1--P2: Pre-training evaluation.} P1 and P2 establish pre-training EDSR on prism and cube grasp-and-lift, respectively, without \AR\ guidance.
    \item \textbf{P3--P5: Training.} Target participants practise 30 prism demonstrations per phase with \AR\ guidance, while Control participants perform 30 matched demonstrations per phase without \AR\ guidance.
    \item \textbf{P6: Trained-task post-test.} Participants repeat prism grasp-and-lift without guidance. A larger P1-to-P6 EDSR improvement in Target than Control would support \(h_2\), beyond the effect of matched unguided practice.
    \item \textbf{P7: Object-transfer post-test.} Participants repeat cube grasp-and-lift without guidance. A larger P2-to-P7 EDSR improvement in Target than Control would support object transfer (\(h_3\)), since cube is encountered at baseline but excluded from guided training.
    \item \textbf{P8: Task-transfer test.} Participants perform ordered pick-and-place of a parallelepiped and a pentagram-shaped object into a bowl without guidance. Higher Target EDSR would support task transfer (\(h_3\)) to this longer-horizon composition.
\end{itemize}
P3--P5 are used only for participant training without data collection; robot policies are trained and evaluated using newly collected demonstrations from P1, P2, and P6--P8.
Removing guidance in all post-training evaluations tests the acquired teaching behaviour.

For Target training, corridor geometry, contact faces, orientation tolerance, reference speed, and thresholds are specified by experts for the robot and workspace.
The corridor radii are \(25/5/3\,\mathrm{cm}\).
The opposing contact faces are selected from a stable physical Franka grasp, with an orientation tolerance of \(15^\circ\) from vertically downward.
The task-normalised speed cue is green for \(v\leq2v_{\mathrm{ref}}\) and red otherwise throughout the trajectory.

\begin{figure}[!htbp]
    \centering
    \includegraphics[width=\columnwidth]{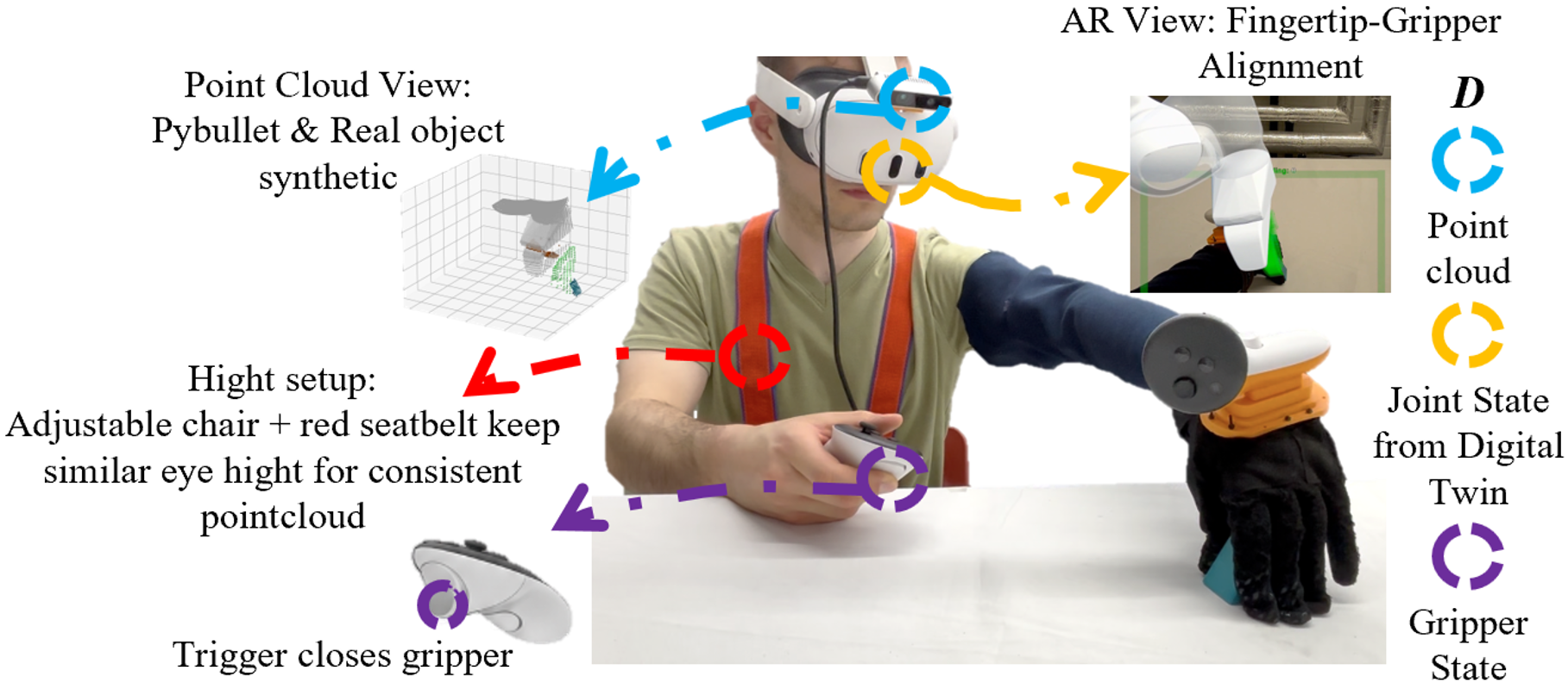}

    \caption{Shared demonstration-collection interface based on ARCap.
    Quest 3 displays the digital twin alongside the real-object point cloud, with the virtual gripper aligned to the participant's fingertips.
    IK joint actions and trigger-derived gripper states are recorded with the observations; adjusted chair height standardises viewpoints.}
    \label{fig:data_collection_system}
    \vspace{-2.5mm}
\end{figure}

Participants manipulate real objects by hand while aligning their fingertips with the virtual gripper of a PyBullet Franka digital twin (Fig.~\ref{fig:data_collection_system}).
Recorded observations combine real RGB--D scene points with virtual gripper points; timestamped frames pair these observations with inverse-kinematics joint actions and trigger-derived gripper states.
This collection pipeline is shared across groups.

For each participant and evaluation phase, a separate policy is trained from the corresponding 30-demonstration dataset and evaluated over \(K=10\) physical-robot rollouts.
A rollout succeeds only if the prescribed task is completed without intervention, object loss, unsafe contact, collision, or workspace-limit violation.
Before learning, all demonstrations are verified as task-successful by replay on the physical Franka with the Fin-Ray gripper.

\subsection{Retained Transfer After Two Months}
\label{s:exp_followup}

Immediate improvement does not establish that teaching skill remains useful after a delay.
The follow-up therefore tests \(h_4\) through delayed transfer to tasks and objects outside the original curriculum (\(h_3\)).
The same Target (\(n_p=10\)) and Control (\(n_p=10\)) participants return approximately two months later, having had no access to the study apparatus during the interval.
They receive no refresher training and no \AR\ guidance during either follow-up task.
Higher Target EDSR on these new tasks would support retained transfer of teaching skill, not retention of a previously learned robot policy, since every policy is trained afresh.

\begin{figure}[!htbp]
    \centering
    \includegraphics[width=\columnwidth]{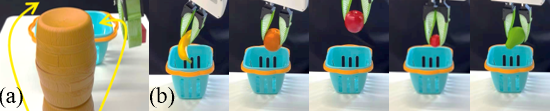}

    \caption{Two-month follow-up tasks conducted without refresher training or AR guidance.
    (a) Bimodal obstacle task with two protocol-prescribed valid routes.
    (b) Fruit tasks with banana, orange, apple, strawberry, and pear.
    Each participant collects new data for policy training.}
    \label{fig:followup_tasks}
    \vspace{-2.5mm}
\end{figure}

The first task asks whether the acquired skill accommodates multiple valid solutions rather than enforcing a single path: obstacle-avoidance grasping (Fig.~\ref{fig:followup_tasks}(a)).
Motion over the obstacle is disallowed; within the bounded workspace, this leaves two feasible route classes, passing the obstacle on the left (L) or right (R).
Each participant collects 30 demonstrations per route (L and R). Route use records whether deployment follows L or R; conditional deployment success \(\hat J_{\mathrm L}\) and \(\hat J_{\mathrm R}\) is computed within the selected route.
Total EDSR tests the teaching advantage; route use and conditional EDSR assess whether both solutions remain represented and whether the advantage holds within each route.

The subsequent fruit task tests whether the delayed teaching advantage extends beyond the block objects used in the original study.
Participants collect 30 demonstrations for each of five fruits: banana, orange, apple, strawberry, and pear (Fig.~\ref{fig:followup_tasks}(b)).
A participant-level policy is trained for every fruit and evaluated over \(K=10\) rollouts.

\subsection{Independent Teaching versus Online AR Feedback}
\label{s:exp_arcap}

Complementing the unguided Control comparison for \(h_2\) and \(h_3\), this experiment asks whether independently applied teaching skill offers an advantage over generic feedback available during collection.
A separate cohort of \(n_p=10\) novices uses ARCap-no-glove~\cite{chen2025arcap} as an online-feedback baseline.
These participants complete the same P1--P5 tasks and demonstration counts as the Target group, using the full ARCap-no-glove feedback configuration, including view, speed, collision, and kinematic-feasibility warnings.
The feedback remains active during collection of 30 demonstrations for each P6--P8 task: prism, cube, and ordered pick-and-place.
The policy comparison uses the same learner, dataset budget, and evaluation protocol for both groups: Target collects after DEMT guidance is removed, whereas ARCap provides online support during collection.
Higher Target EDSR would favour pre-collection learner-aware training over this online-feedback configuration, but would not isolate the effect of guidance timing because the feedback content also differs.
The ARCap participants do not take part in the two-month follow-up.

\subsection{Evaluation Metrics}
\label{s:metrics}

All hypothesis tests evaluate teaching through the deployment performance of a newly learned policy.
The primary metric is empirical deployment success rate (EDSR):
\begin{equation}
\label{e:edsr}
\hat{J}\!\left(\learner(\dataset_z)\right)
=
\frac{1}{K}\sum_{k=1}^{K}\successf(\tau_k),
\end{equation}
where \(\successf(\tau_k)\in\{0,1\}\) indicates whether rollout \(k\) satisfies the task and safety criteria.
Human-study statistical tests use participant-level EDSR, with pre-to-post changes testing improvement and between-group comparisons on transfer tasks testing the teaching advantage beyond the curriculum.

To examine whether deployment gains accompany changes in the targeted demonstration structures, three secondary diagnostics are computed after normalising each trajectory to \(T=100\) samples.
Mean path coordinate variance (MPCV) measures spatial consistency:
\begin{equation}
\label{e:mpcv}
\operatorname{MPCV}
=
\frac{1}{T}\sum_{t=1}^{T}
\operatorname{Var}_{i}\!\left(\path_t^{(i)}\right).
\end{equation}
Mean closing-orientation dispersion (MCOD) is
\begin{equation}
\label{e:mcod}
\operatorname{MCOD}
=
\frac{1}{|\mathcal{W}|}
\sum_{\tau\in\mathcal{W}}
\left\langle d_q(q_{i,\tau},q_{j,\tau})\right\rangle_{i<j},
\end{equation}
where \(q_{i,\tau}\equiv q_{c_i+\tau}^{(i)}\) and
\(d_q(q_a,q_b)\equiv 2\arccos\!\left(|q_a^\top q_b|\right)\).
Mean path-progress synchronisation variance (MPSV) measures temporal consistency:
\begin{equation}
\label{e:mpsv}
\begin{aligned}
\operatorname{MPSV}
&=
\frac{1}{T}\sum_{t=1}^{T}
\operatorname{Var}_{i}\!\left(\rho_t^{(i)}\right),\\
\rho_t^{(i)}
&=
\frac{\sum_{k=1}^{t-1}\ell_k^{(i)}}
     {\sum_{k=1}^{T-1}\ell_k^{(i)}}.
\end{aligned}
\end{equation}
Here, \(\ell_k^{(i)}\equiv\|\path_{k+1}^{(i)}-\path_k^{(i)}\|_2\), \(i,j\) index demonstrations, \(c_i\) is the closing index, \(\mathcal W\) is a fixed closing-event window, and \(\rho_t^{(i)}\) is completed path fraction.
Lower MPCV, MCOD, and MPSV indicate lower spatial, contact-orientation, and temporal variation, respectively.
Learner fit is diagnosed by the fixed Diffusion Policy denoising loss over
branches \(\mathcal B=\{\mathrm{arm},\mathrm{grip}\}\):
\begin{equation}
\label{e:val_loss}
\Loss
 =
\sum_{b\in\mathcal B}\lambda_b
\left\|
\hat{\boldsymbol{\epsilon}}_{\theta}^{\,b}
-\boldsymbol{\epsilon}^{b}
\right\|_2^2.
\end{equation}
These are secondary diagnostics: MPCV, MCOD, and MPSV quantify the demonstration structures targeted by the curriculum, while validation loss quantifies fixed-learner fit ($(\lambda_{\mathrm{arm}},\lambda_{\mathrm{grip}})=(0.7,0.3)$); EDSR is the criterion used to establish deployment learnability.

\section{Results}
\label{s:results}

\begin{table}[!htbp]
\centering
\scriptsize
\setlength{\tabcolsep}{2.5pt}
\begin{tabular}{llcc}
\toprule
& & \multicolumn{2}{c}{\(\hat J\) (\%)}\\
\cmidrule(lr){3-4}
Type & Condition & Grasping & Insertion\\
\midrule
\textbf{Best} & \textbf{Consistent}(\(\boldsymbol{z_0}\)) & \(\boldsymbol{87.5\pm9.6}\) & \(\boldsymbol{62.5\pm9.6}\)\\
\cmidrule(lr){1-4}
\multirow{3}{*}{Variant}
 & Spatial(\(z_{\mathcal P}\))    & \(10.0\pm8.2\) & \(12.5\pm5.0\)\\
 & Orientation(\(z_{\mathcal R}\)) & \(2.5\pm5.0\)  & \(2.5\pm5.0\)\\
 & Temporal(\(z_{\mathcal V}\))    & \(2.5\pm5.0\)  & \(0.0\pm0.0\)\\
\bottomrule
\end{tabular}

\caption{Deployment of policies trained from task-successful simulated datasets. Values are mean \(\pm\) s.d. over four policies; each uses \(N=30\) demonstrations and \(K=10\) rollouts.}
\label{tab:expA_results}
\vspace{-2.5mm}
\end{table}

\begin{table*}[!t]
\centering
\scriptsize
\setlength{\tabcolsep}{3.5pt}
\renewcommand{\arraystretch}{1.10}
\begin{tabular*}{\textwidth}{@{\extracolsep{\fill}}lllcccc@{}}
\toprule
Task & Group & Phase
& MPCV (\(\mathrm{m}^2\))
& MCOD (rad)
& MPSV
& Val. \(\Loss\)\\
\midrule
\multirow{4}{*}{Prism}
 & \multirow{2}{*}{Target}  & P1 & \(0.0046{\pm}0.0016\) & \(0.3342{\pm}0.0709\) & \(0.0026{\pm}0.0004\) & \(0.2646{\pm}0.1240\)\\
 &                          & P6 & \(\mathbf{0.0007{\pm}0.0004}\) & \(\mathbf{0.0749{\pm}0.0051}\) & \(\mathbf{0.0013{\pm}0.0003}\) & \(\mathbf{0.0710{\pm}0.0230}\)\\
\cmidrule(lr){2-7}
 & \multirow{2}{*}{Control} & P1 & \(0.0052{\pm}0.0009\) & \(0.3732{\pm}0.0382\) & \(0.0028{\pm}0.0004\) & \(0.2316{\pm}0.0653\)\\
 &                          & P6 & \(0.0053{\pm}0.0007\) & \(0.3757{\pm}0.0228\) & \(0.0031{\pm}0.0002\) & \(0.2252{\pm}0.0886\)\\
\midrule
\multirow{4}{*}{Cube}
 & \multirow{2}{*}{Target}  & P2 & \(0.0083{\pm}0.0014\) & \(0.6582{\pm}0.1383\) & \(0.0056{\pm}0.0017\) & \(0.2460{\pm}0.0650\)\\
 &                          & P7 & \(\mathbf{0.0005{\pm}0.0003}\) & \(\mathbf{0.1668{\pm}0.0296}\) & \(\mathbf{0.0010{\pm}0.0002}\) & \(\mathbf{0.0606{\pm}0.0084}\)\\
\cmidrule(lr){2-7}
 & \multirow{2}{*}{Control} & P2 & \(0.0086{\pm}0.0008\) & \(0.6806{\pm}0.0632\) & \(0.0058{\pm}0.0016\) & \(0.2570{\pm}0.0733\)\\
 &                          & P7 & \(0.0088{\pm}0.0011\) & \(0.6809{\pm}0.0623\) & \(0.0055{\pm}0.0028\) & \(0.2295{\pm}0.0676\)\\
\bottomrule
\end{tabular*}

\caption{Data structure and learner fit without AR guidance (mean \(\pm\) s.d., 10 participant-level datasets/group).
P1/P2 are pre-training and P6/P7 post-training; lower is better.
P3--P5 are training-only; complete P8 and ARCap diagnostics appear in the supplement video.}
\label{tab:prepost_structure}
\vspace{-4.5mm}
\end{table*}

\begin{figure*}[!t]
    \centering
    \includegraphics[width=\textwidth]{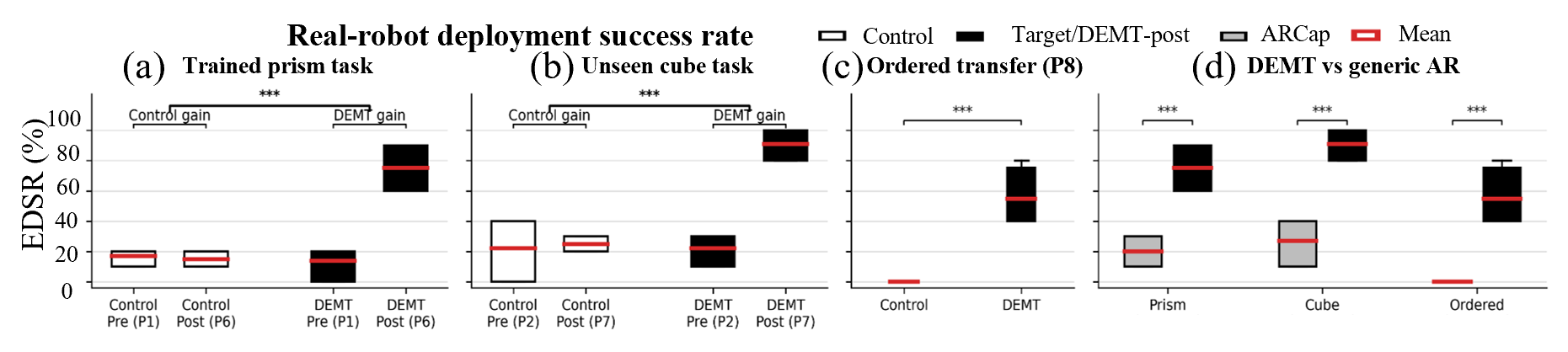}

    \caption{Immediate real-robot EDSR.
    (a) Pre-to-post performance on the trained prism task (P1 to P6).
    (b) Pre-to-post transfer to the cube held out from AR-guided training (P2 to P7).
    (c) Transfer to the ordered composition (P8).
    (d) Post-training DEMT performance without AR guidance versus the generic online AR baseline on the same P6--P8 tasks.
    Target and Control policies use datasets collected without AR guidance; ARCap policies in (d) use datasets collected with generic online feedback.
    Rectangles span the 0th--90th percentiles of participant-level EDSR; red lines denote group means; \(^{***}p<0.001\).}
    \label{fig:immediate_results}
    \vspace{-4.5mm}
\end{figure*}

\begin{figure}[!htbp]
    \centering
    \includegraphics[width=\columnwidth]{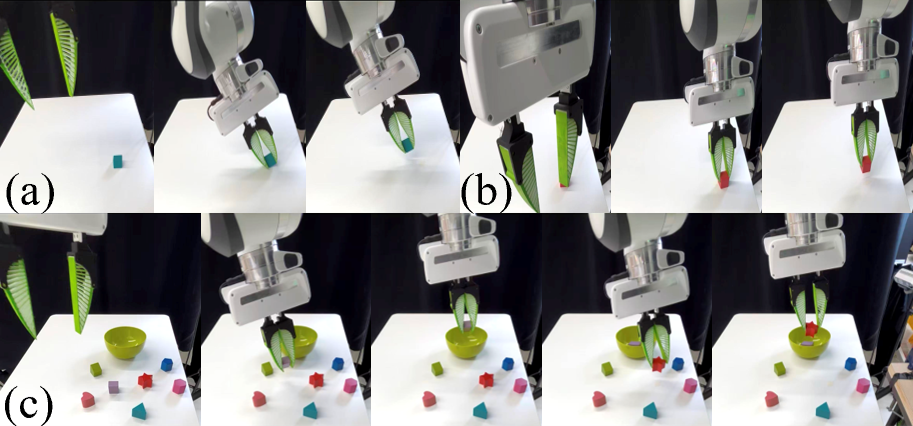}

    \caption{Representative successful physical-robot rollouts for the immediate evaluation tasks; frames proceed from left to right.
    (a) Prism grasp-and-lift.
    (b) Cube grasp-and-lift, with the cube held out from the AR-guided curriculum.
    (c) Related ordered two-object composition.
    AR guidance is absent during these evaluations, and each task uses a separate policy trained from the corresponding dataset.}
    \label{fig:immediate_rollouts}
    \vspace{-4.5mm}
\end{figure}

\subsection{Task Success Does Not Determine Learnability ($h_1$)}
\label{s:results_sim}

Although every scripted demonstration succeeds, the resulting policies differ sharply (Table~\ref{tab:expA_results}).
Relative to \(z_0\), the three isolated variants(\(z_{\mathcal P}\), \(z_{\mathcal R}\), and \(z_{\mathcal V}\)) reduce deployment success by 70 percentage points on average.
All six variant--baseline comparisons are significant under two-sided Welch tests (\(p<0.001\)).
Within the finite set, \(\hat z=z_0\); its PRV structures provide the targets translated into the current curriculum.

\subsection{AR Guidance Improves Subsequent Teaching ($h_2,h_3$)}
\label{s:results_immediate}

Before training, Target and Control do not differ significantly on either prism (P1, \(p=0.708\); Fig.~\ref{fig:immediate_results}(a)) or cube (P2, \(p=0.844\); Fig.~\ref{fig:immediate_results}(b)).
On prism, Target improves from \(14.0\pm9.7\%\) in P1 to \(75.0\pm10.8\%\) in P6, conducted without \AR\ guidance (Fig.~\ref{fig:immediate_results}(a)).
Control changes from \(17.0\pm4.8\%\) to \(15.0\pm5.3\%\).
The between-group difference in improvement is significant (two-sided Mann--Whitney, \(U=100\), \(p<0.001\)).
Because P6 uses newly collected data and a newly trained policy, with no \AR\ guidance during collection or evaluation, the gain reflects changed demonstrator behaviour.

The effect transfers to cube: Target rises from \(22.0\pm7.9\%\) in P2 to \(91.0\pm8.8\%\) in P7, while Control changes from \(22.0\pm14.8\%\) to \(25.0\pm5.3\%\) (\(U=100\), \(p<0.001\); Fig.~\ref{fig:immediate_results}(b)).
On the related longer-horizon P8 composition, Target achieves \(55.0\pm12.7\%\) and Control \(0.0\pm0.0\%\) (\(U=100\), \(p<0.001\); Fig.~\ref{fig:immediate_results}(c)).
Representative successful physical-robot rollouts are shown in Fig.~\ref{fig:immediate_rollouts}.

The data-level measures (MPCV, MCOD, and MPSV) and learner-level validation loss show the same overall trend as EDSR: Target improves after training and outperforms Control (Table~\ref{tab:prepost_structure}).
Together, these results support improved teaching on the trained task (\(h_2\)) and immediate transfer beyond the guided curriculum (\(h_3\)).
The cube tests object transfer, while the ordered composition extends the acquired skill to a longer-horizon task.

\begin{table}[!htbp]
\centering
\scriptsize
\textbf{(a) Bimodal obstacle task}\\[2pt]
\setlength{\tabcolsep}{2.5pt}
\begin{tabular}{llcc}
\toprule
Measure & Metric & Target & Control\\
\midrule
\multirow{2}{*}{Route use}
 & L (\%) & \(42.5\pm14.4\) & \(43.0\pm10.3\)\\
 & R (\%) & \(57.5\pm14.4\) & \(57.0\pm10.3\)\\
\cmidrule(lr){1-4}
\multirow{3}{*}{EDSR}
 & \(\hat J_{\mathrm L}\) (\%) & \(\boldsymbol{52.2\pm13.2}\) & \(27.0\pm23.8\)\\
 & \(\hat J_{\mathrm R}\) (\%) & \(\boldsymbol{43.5\pm21.3}\) & \(23.8\pm14.3\)\\
 & Total \(\hat J\) (\%)       & \(\boldsymbol{48.5\pm12.3}\) & \(24.5\pm13.4\)\\
\bottomrule
\end{tabular}
\par
\scriptsize
\textbf{(b) Fruits}\\[2pt]
\setlength{\tabcolsep}{5pt}

\begin{tabular}{lcc}
\toprule
Object & Target \(\hat J\) (\%) & Control \(\hat J\) (\%)\\
\midrule
Banana     & \(\boldsymbol{93\pm8.2}\)  & \(60\pm11.5\)\\
Orange     & \(\boldsymbol{87\pm9.5}\)  & \(45\pm10.8\)\\
Apple      & \(\boldsymbol{81\pm9.9}\)  & \(38\pm11.4\)\\
Strawberry & \(\boldsymbol{72\pm11.4}\) & \(32\pm13.2\)\\
Pear       & \(\boldsymbol{90\pm8.2}\)  & \(53\pm9.5\)\\
\bottomrule
\end{tabular}

\caption{Two-month follow-up without AR guidance.
Values are mean \(\pm\) s.d. over 10 participant-level policies per group.
The bimodal obstacle policy uses the shuffled union of \(30_{\mathrm L}+30_{\mathrm R}\) prescribed-route demonstrations and \(K=20\) rollouts; each subsequent fruit policy uses 30 new demonstrations collected without AR guidance and \(K=10\) rollouts.}
\label{tab:followup_results}
\vspace{-5.5mm}
\end{table}
\subsection{Retained Transfer Across Tasks and Objects ($h_3,h_4$)}
\label{s:results_obstacle}
\label{s:results_fruit}

At the two-month follow-up, Target and Control deployments express similar route proportions on the bimodal obstacle task, with both prescribed solutions represented (Table~\ref{tab:followup_results}).
Target nevertheless achieves significantly higher total EDSR than Control (\(48.5\%\) versus \(24.5\%\); \(U=90\), \(p=0.0026\)).
Conditional performance is also higher for Target along both routes.

In the subsequent fruit evaluation, Target policies outperform Control across all five fruits, averaging \(84.6\%\) versus \(45.6\%\) EDSR (Table~\ref{tab:followup_results}).
All five per-fruit comparisons are significant (all \(p<0.001\)).
Together, these results support transfer of the acquired teaching skill to a task with multiple valid solutions and to previously untrained objects (\(h_3\)).
Because these transfers are observed two months after training without refresher guidance, they also support \(h_4\).

\subsection{Comparison with Online AR Feedback}
\label{s:results_arcap}

After matched task exposure in P1--P5, policies trained from demonstrations collected with ARCap-no-glove achieve \(20.0\pm8.2\%\), \(27.0\pm12.5\%\), and \(0.0\pm0.0\%\) EDSR on prism, cube, and ordered pick-and-place, respectively (Fig.~\ref{fig:immediate_results}(d)).
The corresponding Target policies, trained from post-\DEMT\ demonstrations collected without \AR\ guidance, achieve \(75.0\pm10.8\%\), \(91.0\pm8.8\%\), and \(55.0\pm12.7\%\), respectively (\(U=100\), \(p<0.001\) for all three comparisons; Fig.~\ref{fig:immediate_results}(d)).
Target therefore achieves higher deployment success with demonstrations collected independently after training than the baseline achieves with online feedback during collection.
This comparison supports the value of learner-aware training for subsequent demonstration quality.

\section{Discussion}

At a fixed, empirically sufficient demonstration budget, novices can improve the learning value of their demonstrations and apply the acquired teaching skill after guidance is removed. \DEMT\ connects learner-dependent demonstration quality to human training: deployment evaluation identifies a target structure, and AR cues guide participants in practising it. Higher EDSR is accompanied by lower MPCV, MCOD, MPSV, and validation loss, linking changes in demonstration structure to improved learner fit and robot performance.

Immediate transfer to the cube and ordered composition, followed by the two-month obstacle and fruit results, suggests that participants learned a reusable teaching strategy rather than only a task-specific movement. The obstacle task also clarifies the role of consistency: useful structure need not eliminate diversity. Both routes remained represented in deployment, while Target demonstrations supported higher success within each route.

The ARCap comparison places this independent teaching ability in context. After matched task exposure, Target demonstrations collected without training cues yield higher deployment success than those collected with generic online feedback. The benefit of DEMT thus extends beyond assisted practice to subsequent data collection.

\textbf{Limitations:} The study uses an expert-defined candidate set and AR cues, one learner and robot platform, static tabletop tasks, and \(n_p=10\) participants per group. Future work should automate candidate and guidance generation, broaden the human-controllable teaching space, and examine transfer across learners, embodiments, larger cohorts, and more dexterous tasks.

\section{Conclusion}
\label{s:conclusion}

\DEMT\ uses robot deployment to select demonstration structures and translates them into AR training for novice demonstrators.
After guidance removal, trained participants provide more learnable demonstrations on the trained prism, a cube held out from the guided curriculum, and an ordered composition.
The teaching advantage extends to a bimodal obstacle task and five fruits two months later without refresher training.
These findings support training human demonstrators as a means of improving subsequent robot learning, with transferable teaching skills that accommodate structured diversity.

\addtolength{\textheight}{0cm}   

{\scriptsize\printbibliography}%
\end{document}

%% file: tex/preamble.tex
\usepackage{times}
\usepackage{graphicx}\graphicspath{{img/}}
\usepackage{epstopdf}
\usepackage{algorithm}
\usepackage{algpseudocode}
\usepackage{multirow} 
\usepackage{multicol}
\usepackage{array}
\usepackage{mdwmath}
\usepackage{mdwtab}
\usepackage{rotating}
\usepackage[list=off]{caption}
\usepackage{amssymb}
\usepackage{verbatim}
\usepackage{overpic}
\usepackage{amsmath}
\usepackage{subcaption}
\usepackage[utf8]{inputenc}
\usepackage[T1]{fontenc}
\usepackage[export]{adjustbox}
\usepackage[normalem]{ulem}
\usepackage{siunitx} 
\usepackage{chngpage}
\usepackage{xcolor}

\input{./tex/abbreviations}
\input{./tex/symbols}
\input{./tex/format}

\makeatletter\newcommand{\manuallabel}[2]{\def\@currentlabel{#2}\label{#1}}\makeatother

\usepackage[
  backend=bibtex,
  style=ieee,
  url=false,
  isbn=false,
  doi=false,
  eprint=false
]{biblatex}
\renewcommand{\cite}[1]{~\autocite{#1}}

%% file: tex/abbreviations.tex
\usepackage[acronym]{glossaries}%
\glsdisablehyper 
\makeglossaries%

\newacronym{AI}{AI}    {artificial intelligence}%
\newacronym{RL}{RL}    {reinforcement learning}%
\newacronym{LfD}{LfD}  {learning from demonstration}\providecommand{\LfD}{\gls{LfD}}%
\newacronym{MT}{MT}    {machine teaching}%
\newacronym{DEMT}{DEMT}    {deployment-evaluated machine teaching}\providecommand{\DEMT}{\gls{DEMT}}%
\newacronym{AR}{AR}    {augmented reality}\providecommand{\AR}{\gls{AR}}%
\newacronym{HCI}{HCI}    {human-computer interaction}%
\newacronym{MPCV}{MPCV}    {mean pointwise Cartesian variance}%
\newacronym{MCOD}{MCOD}    {mean closing orientation dispersion}%
\newacronym{MPSV}{MPSV}    {mean path-progress synchronisation variance}%
\newacronym{EDSR}{EDSR}    {empirical deployment success rate}%

%% file: tex/symbols.tex
\usepackage{amssymb}
\usepackage{mathtools}
\usepackage{amsmath}
\usepackage{gensymb}
\usepackage{algorithm}
\usepackage{algpseudocode}
\usepackage{amsfonts}
\usepackage{bm}

\mathchardef\mhyphen="2D   
\providecommand{\triskf}  {\mathcal{\rho}}    

\providecommand{\terrorf}  {\mathcal{\varepsilon}}    

\providecommand{\obsspace}{\mathbb{O}}  
\providecommand{\obsc}{\mathbf{O}} 

\providecommand{\actionspace}{\mathbb{A}} 
\providecommand{\actionc}{\mathbf{A}} 

\providecommand{\learner}{\mathcal{A}}

\providecommand{\datasetspace}{\mathbb{D}} 
\providecommand{\dataset}{\bm{\mathcal{D}}} 
\providecommand{\Nd}{N} 
\providecommand{\optimaldataset}{\dataset^{*}}

\providecommand{\path}{\mathbf{p}}  

\providecommand{\model}{\bm{\theta}}  
\providecommand{\learntmodel}{\tilde{\model}}  
\providecommand{\targetmodel}{\model^{*}}  

\providecommand{\successf}{\mathcal{S}}
\providecommand{\Loss} {\mathcal{L}}

%% file: tex/format.tex
\let\labelindent\relax 
\usepackage[inline]{enumitem} 
\setlist{nolistsep}
\usepackage[singlelinecheck=off]{caption}